# Conversation as Action Under Uncertainty

**Tim Paek**
Department of Psychology
Stanford University
Stanford, CA 94305
paek@psych.stanford.edu

**Eric Horvitz**
Microsoft Research
One Microsoft Way
Redmond, WA 98052
horvitz@microsoft.com

## Abstract

Conversations abound with uncertainties of various kinds. Treating conversation as inference and decision making under uncertainty, we propose a task independent, multimodal architecture for supporting robust continuous spoken dialog called *Quartet*. We introduce four interdependent levels of analysis, and describe representations, inference procedures, and decision strategies for managing uncertainties within and between the levels. We highlight the approach by reviewing interactions between a user and two spoken dialog systems developed using the *Quartet* architecture: *Presenter*, a prototype system for navigating Microsoft PowerPoint presentations, and the *Bayesian Receptionist*, a prototype system for dealing with tasks typically handled by front desk receptionists at the Microsoft corporate campus.

## 1 INTRODUCTION

Conversations abound with uncertainties of various kinds that may lead to misunderstanding and other communication failures. These uncertainties permeate every level of conversation, from attending to what was said and identifying what words were spoken, to understanding the intentions behind the words. While people manage these multiple uncertainties with almost effortless ease, automated dialog systems often break down on account of them. The functionality and effectiveness of spoken dialog systems relies critically on developing a robust, unified architecture for sharing key uncertainties among disparate components such as automatic speech recognition (ASR), natural language parsing, and even computer vision.

Beginning with a brief overview of research in psychology and linguistics on how people resolve their uncertainties, we discuss methods for approaching conversation as inference and decision making under uncertainty at four levels of analysis in a task independent, multimodal architecture called *Quartet*. We describe representations, inference procedures, and decision strategies for managing uncertainties within and between the levels. We highlight the architecture and the robustness of the approach by reviewing interactions between a user and two prototype spoken dialog systems: *Presenter*, a system for navigating Microsoft PowerPoint presentations, and the *Bayesian Receptionist*, a system for dealing with tasks typically handled by front desk receptionists at the Microsoft corporate campus (Horvitz & Paek, 1999; Paek & Horvitz, 1999).

## 2 TOWARD AN ARCHITECTURE FOR CONVERSATION

When people engage in a conversation, they typically do so with the intent of making themselves understood. They need to make sure, as they speak, that the other participants are at the same time attending to, hearing, and understanding what they are saying. Since unresolved uncertainties can result in communication failures, people collaborate to establish and maintain the mutual belief that their utterances have been understood well enough for current purposes (Clark & Schaefer, 1987, 1989).

An example of this process is the provision of feedback. To establish the mutual belief that an utterance has been sufficiently understood, people will typically give feedback of their understanding through head nods or acknowledgements, such as "uh huh" (Goodwin, 1986). On the other hand, if they do not understand, they will attempt to clear up their uncertainties, or take other measures. In short, people coordinate not only what they say but also their beliefs about what they mutually understand.

Researchers in psychology and linguistics have argued that with this kind of elaborate coordination, a conversation is more reminiscent of a collaborative effort or joint activity than simply a structured sequence of utterances (Clark, 1996; Cohen & Levesque, 1991, 1994). The process by which participants elegantly coordinate the presentation and acceptance of their utterances to establish, maintain, and confirm mutual understanding has been called *grounding* (Clark & Brennan, 1991; Clark & Schaefer, 1987, 1989; Clark & Wilkes-Gibbs, 1990). Grounding involves not only the consideration of how key uncertainties depend on each other and influence mutual understanding, but also what decisions to make in light of these uncertainties. In trying to establish mutual understanding, people appear to assess the costs and benefits of pursuing different avenues of grounding; for example, a person may compare the cost of asking someone for clarification immediately to the cost of working out an unresolved misunderstanding later. As researchers have shown, grounding is crucial to circumventing communication failures, the repair of which may be costly both in terms of time and effort (Brennan & Hulteen, 1995; Hirst et al., 1994; Paek & Horvitz, 1999; Traum & Dillenbourg, 1996).

The approach we take in the Quartet architecture is to treat the process of grounding as decision making under uncertainty. We explicitly represent key uncertainties with Bayesian networks, and use local expected utility and value-of-information analyses to identify actions that maximize mutual understanding and bolster grounding. Since the networks and decision policies focus on the basic process of grounding, they generalize across task domains as well as variations in the capabilities for multimodal interaction, as we illustrate later.

The Quartet architecture is motivated by the idea that the design for a spoken dialog system should center on mechanisms for modeling and exploiting uncertainties. Much of the research and development in dialog systems has operated on the tacit assumption that automating conversation entails little more than fine tuning the precision of components such as speech recognition. In contrast, Quartet provides a framework for maintaining a conversation without the luxury of perfect components. Just as people rely on grounding techniques to compensate for additional uncertainties that may arise in conversations with people of impaired language skills, such as imperfect hearing, Quartet enables a dialog system to adapts its strategies based on its beliefs or representations of mutual understanding, and its evaluation of the costs and benefits of taking various grounding measures. Quartet unifies dialog components by allowing them to share uncertainties with each other, as we discuss in the next section.

The Quartet architecture is informed by research in psychology investigating mutual understanding at successive levels of analysis (Clark & Schaefer, 1987, 1989). While researchers have examined the kinds of communication failures that result from lack of grounding at these levels (Brennan & Hulteen, 1995; Dillenbourg et al., 1996), relatively little work has focused on exploiting uncertainties; for example, by quantifying mutual beliefs in terms of probabilities, and then harnessing the probabilities to make optimal decisions. Research on Quartet broadens the scope of previous models of grounding (Traum, 1994; Traum & Dillenbourg, 1996, 1998) by introducing a decision-theoretic perspective; we highlight the efficacy of Bayesian models and expected utility analysis to capture psychological intuitions about the role of uncertainty in the grounding process.

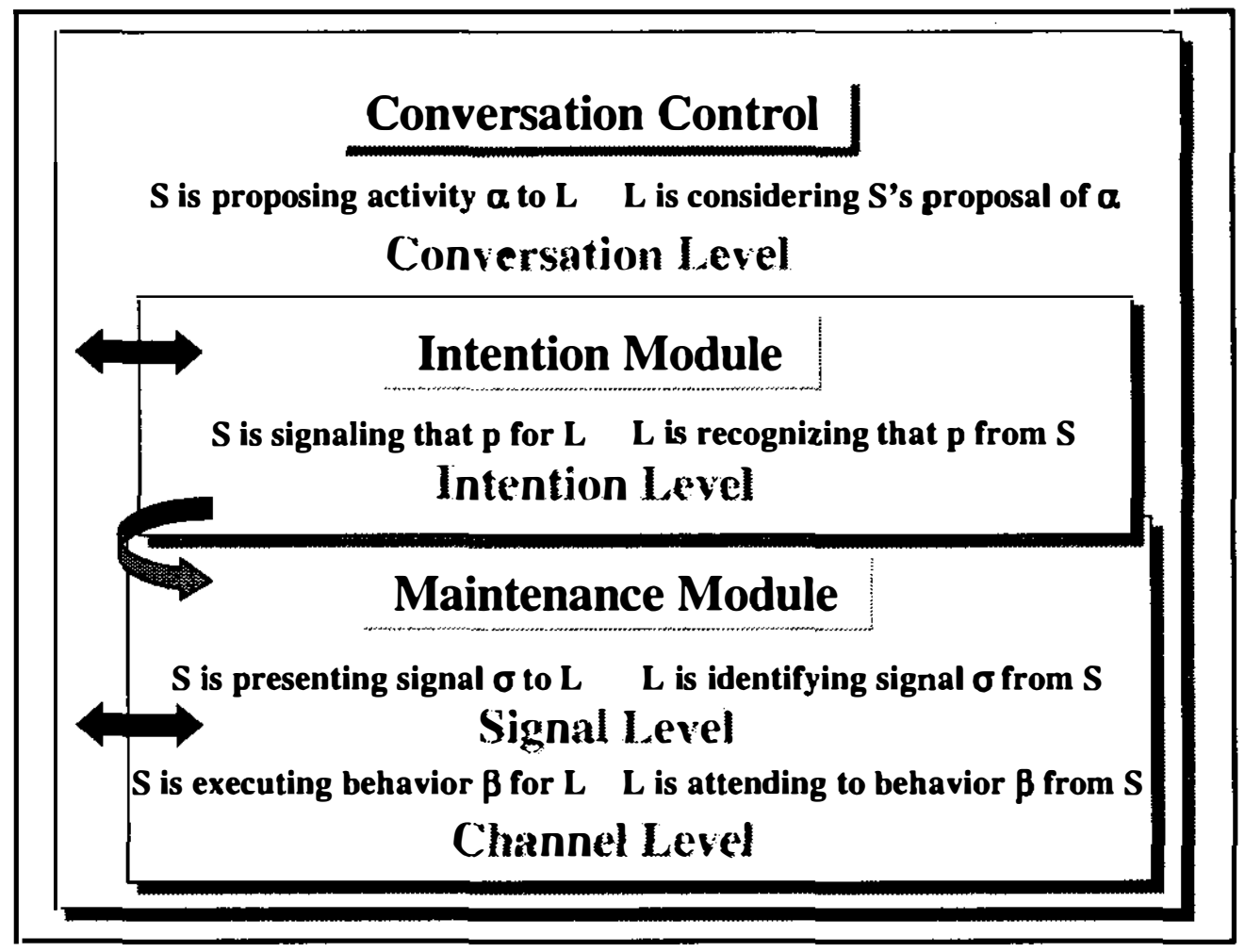

Figure 1. Four levels of representation for inference and decision making under uncertainty in conversation.

Quartet is part of the ongoing *Conversational Architectures* project at Microsoft Research for exploring computational frameworks that capitalize on models of uncertainty to monitor and guide interactive dialog. Using the Bayesian Receptionist as a test domain, Quartet addresses the need for richer, more robust models of how uncertainties affect mutual understanding at multiple levels of conversation. In related work on the *DeepListener* system (Horvitz & Paek, 2000), we have investigated methods for improving spoken command-and-control systems through temporal Bayesian networks that fuse information from multiple utterances. With respect to grounding, we have formulated definitions and decision-theoretic procedures for identifying when *sufficient* mutual understanding has been achieved in a given context (Paek & Horvitz, 2000). Quartet employs ideas and methods developed in these related projects, though we do not discuss them here.

### 2.1 FOUR LEVELS OF ANALYSIS

As mentioned previously, when people engage in a conversation, they make sure that the other participants are at the same time attending to, hearing, and understanding what they are saying. Taking inspiration from Clark (1996), we evaluate grounding at four levels of analysis as displayed in Figure 1.

At the most basic level, which we denote as the *channel level*, a speaker $S$ attempts to open a channel of communication by executing behavior $\beta$, such as an utterance or action, for listener $L$. However, $S$ cannot get $L$ to perceive $\beta$ without coordination; $L$ must be attending to and perceiving $\beta$ *precisely* as $S$ is executing it. Likewise, at the *signal level*, $S$ presents $\beta$ as a signal $\sigma$ to $L$. Not all behaviors are meant to be signals, as for example, scratching an itch. Hence, $S$ and $L$ must coordinate what $S$ presents with what $L$ identifies.

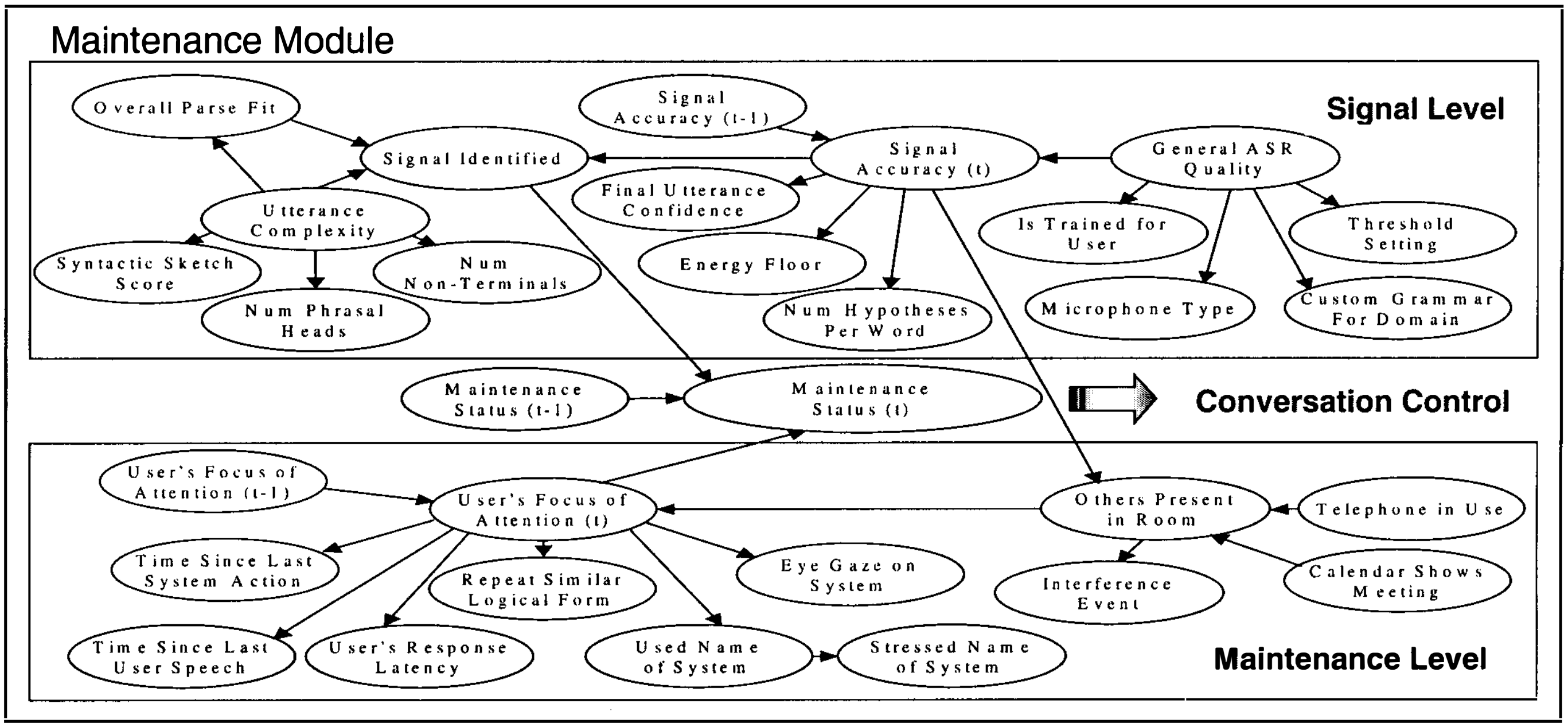

Figure 2. Portions of the temporal Bayesian networks used in the Quartet Maintenance Module.

The *intention level* is where the task of understanding the *semantic content* of signals occurs, and where, to date, efforts on constructing dialog systems have been almost entirely focused. Here, $S$ signals some proposition $p$ for $L$. What $L$ recognizes to be the goal of $S$ in signaling $\sigma$ is *how* $L$ will arrive at $p$. This again takes coordination.

Finally, at the *conversation level*, $S$ proposes some joint activity $\alpha$ which $L$ considers and takes up by providing a conditionally relevant response defined by $\alpha$. $S$ cannot get $L$ to engage in the proposed activity without the coordinated participation and cooperation of $L$.

In summary, all four levels require coordination and collaboration to establish mutual understanding. For spoken dialog systems that integrate diverse components, uncertainties can occur anywhere. Hence, a unified architecture is needed to model probabilistic dependencies between levels. For example, the ASR component at the signal level may pick up words that create confusion for the language understanding component at the intention level. Checking the channel level however, may reveal that the user was actually intending the speech for someone else. Without an architecture that reasons about and exploits uncertainties at these multiple levels, a dialog system would not be able to respond to this typical situation in a robust and intelligent manner.

## 2.2 MODELS, INFERENCE, AND DECISIONS

Quartet uses both atemporal and temporal Bayesian networks (Dagum et al., 1992; Kanazawa & Dean, 1989; Kanazawa et al., 1995; Nicholson & Brady, 1994) to compute the likelihood of variables of interest which we cannot directly observe but only infer as the conversation progresses, such as the overall grounding status. Bayesian networks have been used previously in a number of user modeling projects (Conati et al., 1997; Horvitz, 1997; Horvitz et al., 1998). Before presenting the networks and decision strategies for grounding, we discuss the rationale for the modular design of the architecture.

### 2.2.1 Modularity

As shown in Figure 1, our approach can be viewed as two modules within a larger control subsystem. The *Maintenance Module* handles uncertainties about the channel and signal levels, and the *Intention Module* about the intention level. Surrounding both modules is the *Conversation Control*, which keeps track of the grounding status by continually exchanging information with both modules, as depicted by the arrows. The Conversation Control operates at the meta-level by assessing the status of key variables in all of the modules; it decides where to focus on resolving uncertainties, and what grounding actions to take in light of their likely costs and benefits. By maintaining a historical record of the dialog, including factors such as the number of repair actions taken in prior states of grounding, it can also monitor its own performance and readjust its uncertainties and utilities (Paek & Horvitz, 1999). For example, if the Intention Module continually reports high confidence in knowing the goal of the user even though the Maintenance Module reports cause for concern, and the user repeatedly corrects the dialog system for guessing the wrong goal, the Conversation Control will reassess its internal estimation of the reliability of the Maintenance Module and lower the utility of assuming a goal given the same circumstance.

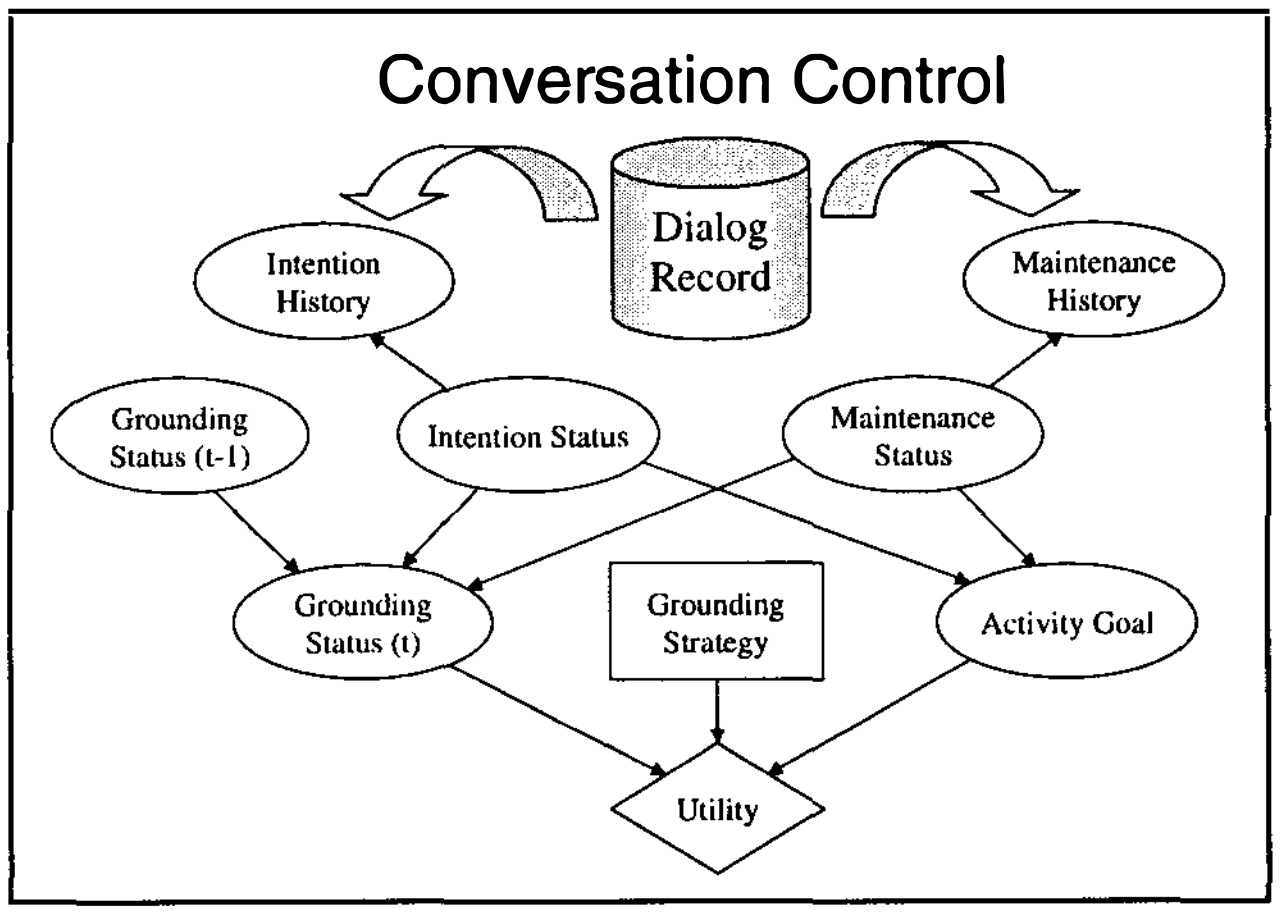

Figure 3. A simple Conversation Control for Quartet.

The design choice of maintaining distinct modules for grounding, as opposed to constructing a single model, was motivated by a number of factors. First, in building Quartet, we found probabilistic dependencies to be abundant within the proposed modules, but sparse between them. This resonates well with the way people focus on inferring and resolving uncertainties at one level before considering other levels; in particular, psychologists have noted that people try to ground understanding at lower levels before moving to higher levels (Clark, 1996). In regards to both psychology and efficiency, building models for distinct modules lends itself to the objective of diagnosis at specific levels. Second, we found that modularity enhances the flexibility of Quartet to be applied to variety of task domains. Since the intention level is where meaning and understanding is coordinated for a particular joint activity, adapting the architecture to new tasks involves nothing more than modifying or exchanging only the Intention Module. The Presenter and Bayesian Receptionist dialog systems employ different Intention Modules while working within the same overall architecture. Quartet also provides an interface to allow other dialog systems to take advantage of grounding information, as most current systems do not employ probabilistic methods. Finally, we found that the beliefs, decisions, and degree of detail necessary for each level varied significantly depending on the communication medium or modality. For example, not all systems have access to vision, or even spoken input. Maintaining modules with more appropriate network structures for different circumstances makes it possible to get the most out of multimodality. Along the same vein, modularity has also been conducive to exploring methods for dynamically switching modules, and even modifying dependencies on the fly with shifts in modality or communication context. For example, if the signal level continually suffers from ASR errors, the system may ask the user to use typed input and then dynamically change its Maintenance Module to reflect this new modality of interaction.

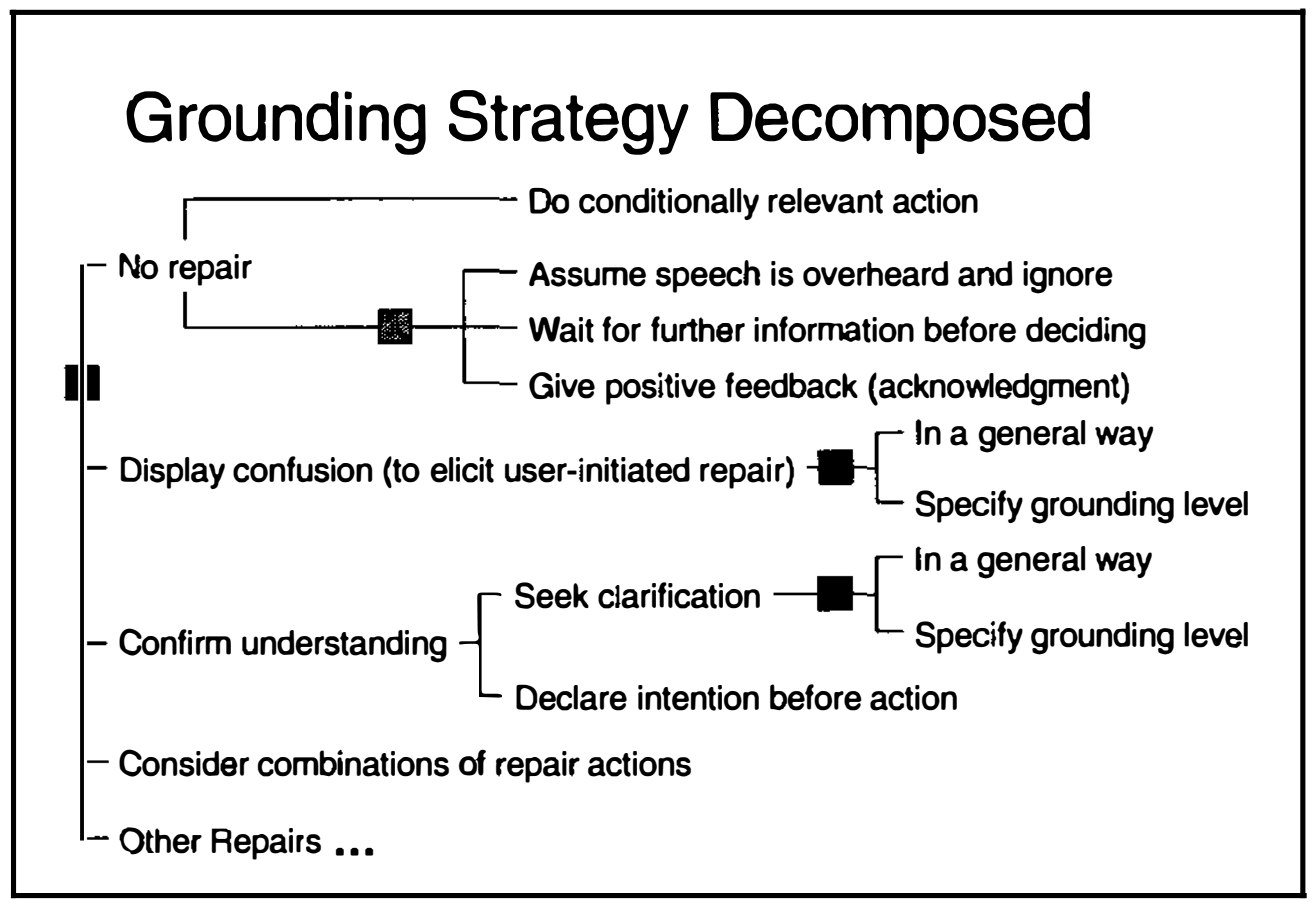

Figure 4. A partial breakdown of possible repair and non-repair grounding strategies in Quartet.

### 2.2.2 Representations for Decision Making

Figure 2 and Figure 3 display portions of the Bayesian models for the Maintenance Module and the Conversation Control respectively. Both networks were initially hand-crafted to take into account variables researchers have identified as being significant in establishing mutual understanding. We have been extending the initial models with user data to learn the parameters and structure of the models.

In the Maintenance Module, beliefs about channel fidelity are captured in the node "User's Focus of Attention," which keeps track of whether the user is attending to the dialog system, another person, or to anything else. For beliefs about the signal level, the node "Signal Identified" summarizes information from the ASR engine and the natural language parser into high, medium, and low confidence states. The "Maintenance Status" node integrates channel fidelity, signal identification, and its own distribution in the previous time slice to obtain a probability distribution over four grounding states: NO CHANNEL, CHANNEL BUT NO SIGNAL, SIGNAL BUT NO CHANNEL, and finally, CHANNEL AND SIGNAL. As we demonstrate later, SIGNAL BUT NO CHANNEL is particularly useful for distinguishing overheard speech from utterances directed at the dialog system.

In the Conversation Control displayed in Figure 3, the "Maintenance Status" node reappears, and is modified to reflect historical performance, as noted in the dialog record. The "Intention Status" node simply conveys how well the "meaning" of the signal was understood, and is also modified by historical performance. "Intention Status" and "Maintenance Status" influence both "Activity Goal," which diagnoses whether the user is participating in a joint activity with the system, another person, or doing something else, and "Grounding Status," which diagnoses the overall mutual understanding in fives states: OKAY, CHANNEL FAILURE, SIGNAL FAILURE, INTENTION FAILURE, and CONVERSATION FAILURE. In order to decide what action to take, the Conversation Control identifies the action with the highest expected utility given the probability distributions of "Grounding Status" and "Activity Goal." Utilities can be elicited from users through psychological experiments (Paek & Horvitz, 1999) or direct assessment tools (Horvitz & Paek, 2000).

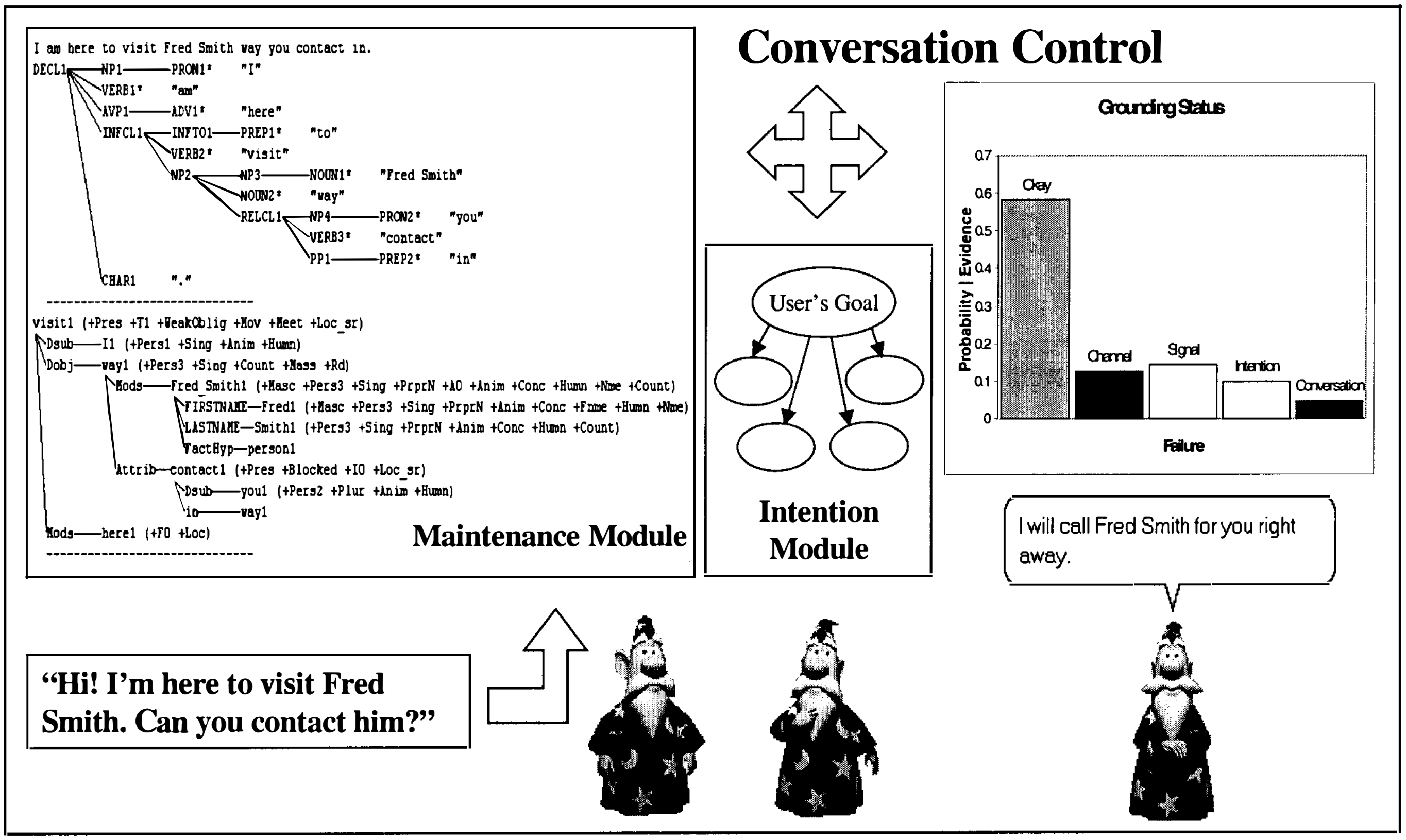

Figure 5. The Bayesian Receptionist using Quartet to check the Grounding Status and provide assistance to users.

Figure 4 shows a partial breakdown of the grounding strategies available in Quartet. Depending on the type of user utterance the system is responding to, Quartet can give feedback of understanding through an acknowledgment, such as "uh huh," or do a "conditionally relevant" action, such as providing a requested service. Repair actions, such as confirming mutual understanding by clarification, can either be general (e.g., "You want a shuttle, right?") or indicative of the specific level of grounding failure (e.g., "I'm not sure if I heard you correctly; did you want a shuttle?"). Furthermore, Quartet is capable of considering the expected utilities of combinations of repair strategies; for example, combining the repair of asking for a repeat with a confirmation produces, "Did you want a shuttle? Can you repeat that?" Utilities for the combinations reflect the fact that not all combinations are natural or desirable.

Figure 3 displays a Conversation Control situation where the expected utility is a function of "Grounding Status" and "Activity Goal." However, we have assessed other Conversation Control modules which decompose the "Grounding Status" into variables such as "User Frustration," "Estimated Time to Finish Joint Activity," and "Perceived Rate of Progress." We have been evaluating the performance of different Conversation Control modules.

### 2.2.3 Refinement through Value of Information

After the Conversation Control selects the grounding strategy with the highest expected utility, it still needs to refine its action by sharing with the user the key uncertainties that affect mutual understanding at a given level of analysis. As psychologists have observed, people "design" their utterances for their audience to improve the chance of mutual understanding (Clark, 1996).

A compelling property of approaching conversation as decision making under uncertainty is the ability to capitalize on value-of-information (VOI) analysis to ask questions, make recommendations, and seek out information in a dialog setting (Horvitz & Paek, 1999). VOI analysis identifies the best evidence to observe in light of the inferred probabilities. To compute VOI, the system calculates for every observation, the expected utility of the best decision associated with each value the observation can take considering the likelihood of different values (Raiffa,1968). Once VOI recommends a piece of evidence to observe, the system can explicitly provide that recommendation or phrase a question to solicit that information. For example, if the grounding strategy is to troubleshoot the signal level for the Maintenance Module in Figure 2, the system may suggest using a different microphone, or ask if other people have suddenly entered the surrounding area.

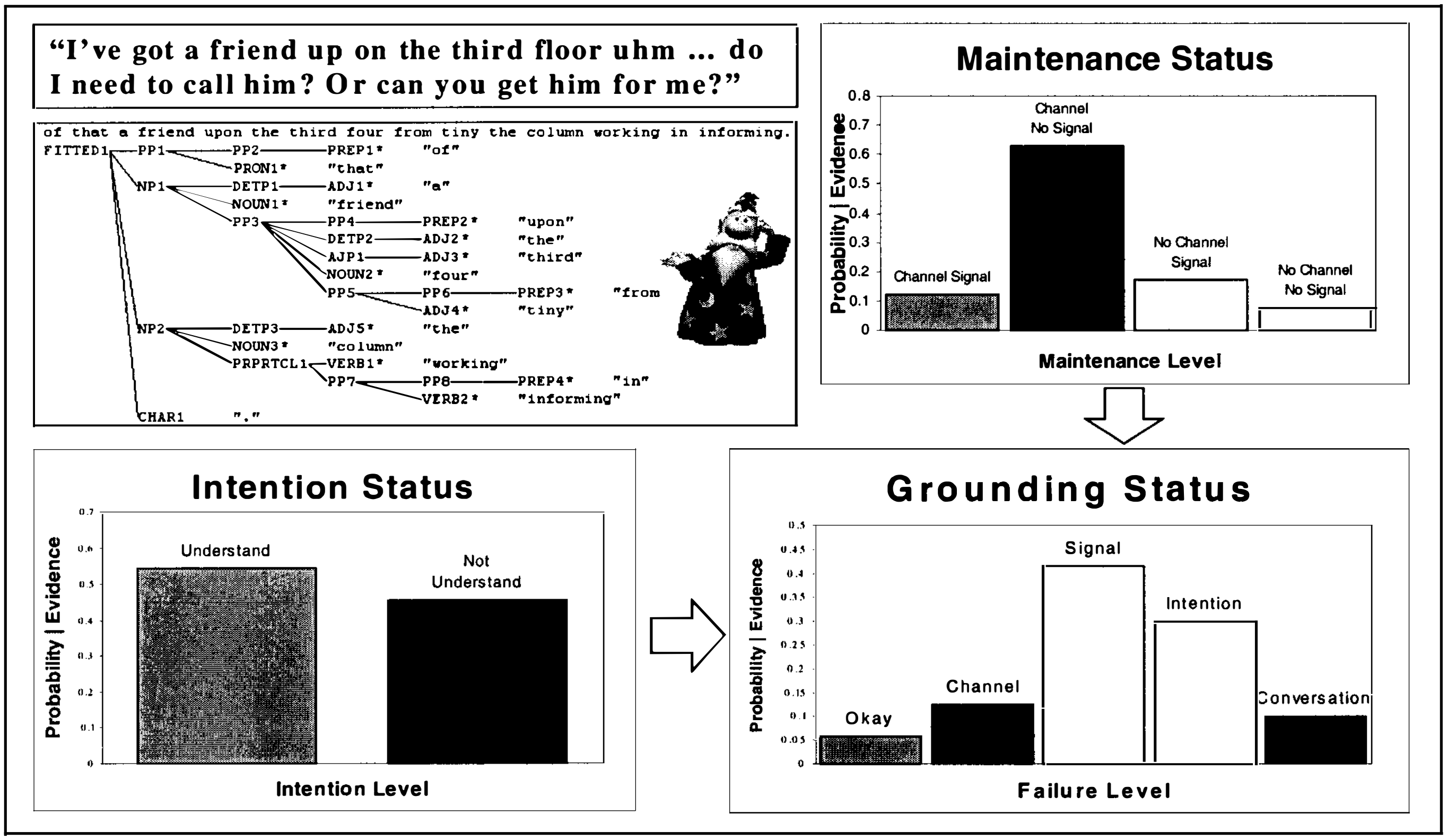

Figure 6. The Receptionist considering uncertainties that cross multiple levels of grounding.

An exact computation of VOI requires the consideration of all possible sequences of observations. However, greedy VOI, focusing on computing the next best piece of evidence to observe, has been found to be a reasonable approximation (Gorry & Barnett, 1968). Within the framework of greedy VOI, a variety of approximations have been employed, including VOI based on minimizing entropy (Ben-Bassat & Teeni, 1984; Heckerman et al., 1992; Horvitz et al., 1988).

# 3 RUNTIME BEHAVIOR

To illustrate the effectiveness of approaching conversation as decision making under uncertainty within a unified architecture that allows components to share uncertainties, we now review interactions between a user and two prototype spoken dialog systems: Presenter, a dialog system for navigating Microsoft PowerPoint presentations, and the Bayesian Receptionist, a dialog system for handling tasks typically addressed by front desk receptionists at the Microsoft corporate campus.

For the following examples, we obtained base ASR information from the Microsoft Whisper system, which contains a trigram language model of about 60,000 words (Huang et al., 1995). No customized grammars were written for either the receptionist or PowerPoint domain. We employed the Microsoft NLPWin natural language parser to obtain syntactic and logical features from recognized speech (Richardson, 1994). These features were instantiated as evidence in the Intention and Maintenance Modules. Finally, we used a face-pose tracker developed at Microsoft Research (Toyama, 1998) for gauging user attention in the Maintenance Module. This computer vision system provides a probability of the user looking directly at the desktop camera.

## 3.1 PROVIDING SERVICES

When all levels of mutual understanding have been sufficiently grounded, Quartet correctly decides that the highest expected utility action is to engage the user in the requested service. The Bayesian Receptionist in Figure 5 presents such an example. Here, a user approaches the Receptionist, looks into the camera, and utters, "***Hi, I'm here to visit Fred Smith. Can you contact him?***" This utterance is recognized by the ASR component (with errors) as "***I am here to visit Fred Smith way you contact in***" which is subsequently parsed by the natural language processing component into the syntactic tree and logical form shown on the left side of Figure 5. Passing syntactic and semantic information to the Intention Module, the most likely goal is "Visitation," which receives a high probability. The probability of this goal is then instantiated as the likelihood of understanding the "meaning" of the utterance in the "Intention Status" node of the Conversation Control. Combining that with the "Maintenance Status," we obtain the distribution for the "Grounding Status" shown on the right side. Given that the "Grounding Status" is most likely OKAY, and that the expected utility of taking action exceeds that of other grounding measures, the Receptionist responds with a predefined template for responding to canonical visitation requests: "***I will call Fred Smith for you right away.***"

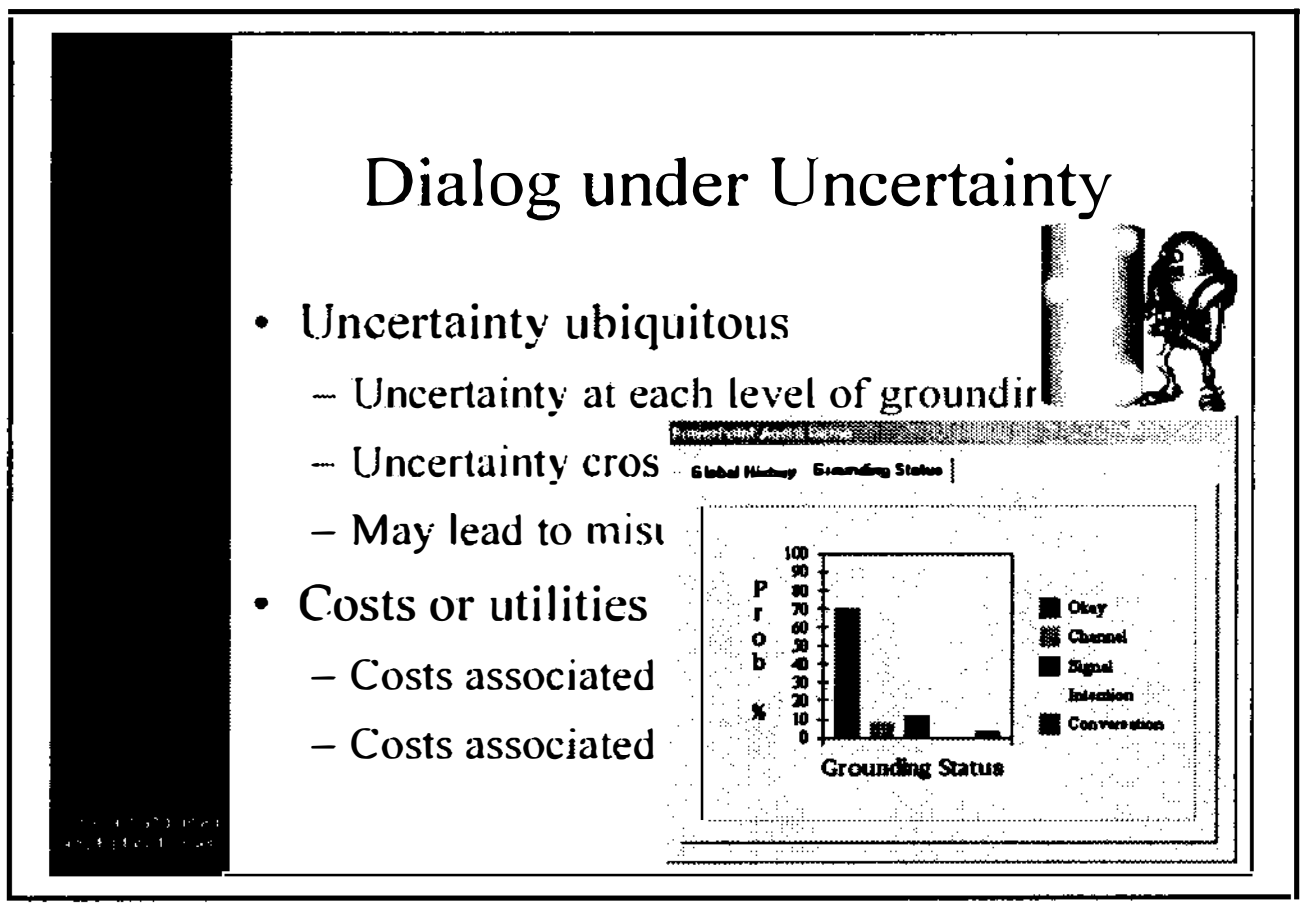

Figure 7. Presenter evaluating the Grounding Status on the fly while moving the presentation to the next slide

### 3.2 EXPLOITING MULTIPLE LEVELS

When the goal of the user at the intention level is not so clear, the system considers uncertainties at lower levels of analysis. Consider a typical example of inter-level dependency. A user approaches the Bayesian Receptionist, looks into the camera, and utters, "***I've got a friend up on the third floor uhm ... do I need to call him? Or can you get him for me?***" The ASR output and subsequent syntactic parse are shown in Figure 6. Such erroneous output is not unusual since ASR engines require a customized grammar for high-quality recognition. Due to the questionable recognition and difficult parsing of the utterance, but at the same time, high degree of attention displayed by the user, the state NO CHANNEL BUT SIGNAL towers in the probability distribution for "Maintenance Status."

At the Conversation Control, the Intention Module reveals that the most likely goal is "Seeking Directions" with a probability of roughly one half. Combining "Intention Status" with "Maintenance Status," we infer the "Grounding Status," which is shown at the bottom of Figure 6. After calculating the expected utilities of various grounding strategies, the action selected is ASK REPEAT, which is appropriate since the highest uncertainty in "Grounding Status" is SIGNAL FAILURE. The final result is that the Receptionist responds to the user by stating, "I'm sorry, I may not have heard you properly. Can you repeat that please?"

Note that by communicating the level of analysis associated with the greatest uncertainty, the system fosters grounding on the part of the users (Brennan & Hulteen, 1995; Paek & Horvitz, 1999). Users are now aware of the need to seek mutual understanding at the signal level. They can therefore design their utterances to make sure that the system correctly recognizes them the next time. If the system continues to suffer SIGNAL FAILURE as the primary "Grounding Status," it may elect to troubleshoot the level using VOI, as discussed previously.

### 3.2 DISTINGUISHING OVERHEARD SPEECH

A major hindrance to rendering spoken dialog systems "hands free," or capable of continuous listening without requiring either a push-to-talk device or the use of a special herald to activate listening, is the problem of distinguishing speech which is intended for the system from that which is overheard. The Quartet architecture is particularly well suited for dealing with this type of grounding problem.

To illustrate how Quartet distinguishes overheard speech, we turn to Presenter, shown navigating a PowerPoint presentation in Figure 7. Here, the user begins by facing the camera and stating, "***Next slide please***," which is recognized by the ASR engine as "***Next slide Luiz.***" Since the most likely goal in the "Intention Status" node, namely NEXT SLIDE, is high and the "Maintenance Status" node indicates that both channel and signal levels are open, the state OKAY dominates "Grounding Status" and Presenter selects providing the requested service as the optimal grounding strategy, moving the presentation forward to the next slide.

Suppose now that the user turns to the audience to continue giving a lecture. Since the microphone is still active, Presenter continually detects words that are well recognized but may or may not be understood at the intention level. Given that the most likely "Maintenance Status" state is SIGNAL BUT NO CHANNEL, if the probability of understanding the goal in "Intention Status" is low, then the expected utility of assuming the speech is overheard will most likely outweigh the expected utility of taking some kind of action, whether it be a repair or non-repair, and the system will ignore the speech. On the other hand, if the probability is high, depending on how well grounding strategies at the intention level have performed in the dialog record, the system may choose to take action, in some cases by simply displaying a sign of confusion. Figure 8 displays a change in expected utilities for two decisions: action and inaction.

Suppose that the probability is low and Presenter ignores the speech. Continuing on with the example, the user now turns back to the camera after a prolonged delay with no action on the part of Presenter, and states "***I want to go back a slide.***" The utterance is recognized as "***Acrobat a side***," and the most likely "Maintenance Status" is CHANNEL BUT NO SIGNAL. This renders ACTION more desirable than INACTION. At this point, the system selects the grounding strategy with the highest expected utility, just as in the example of the previous section.

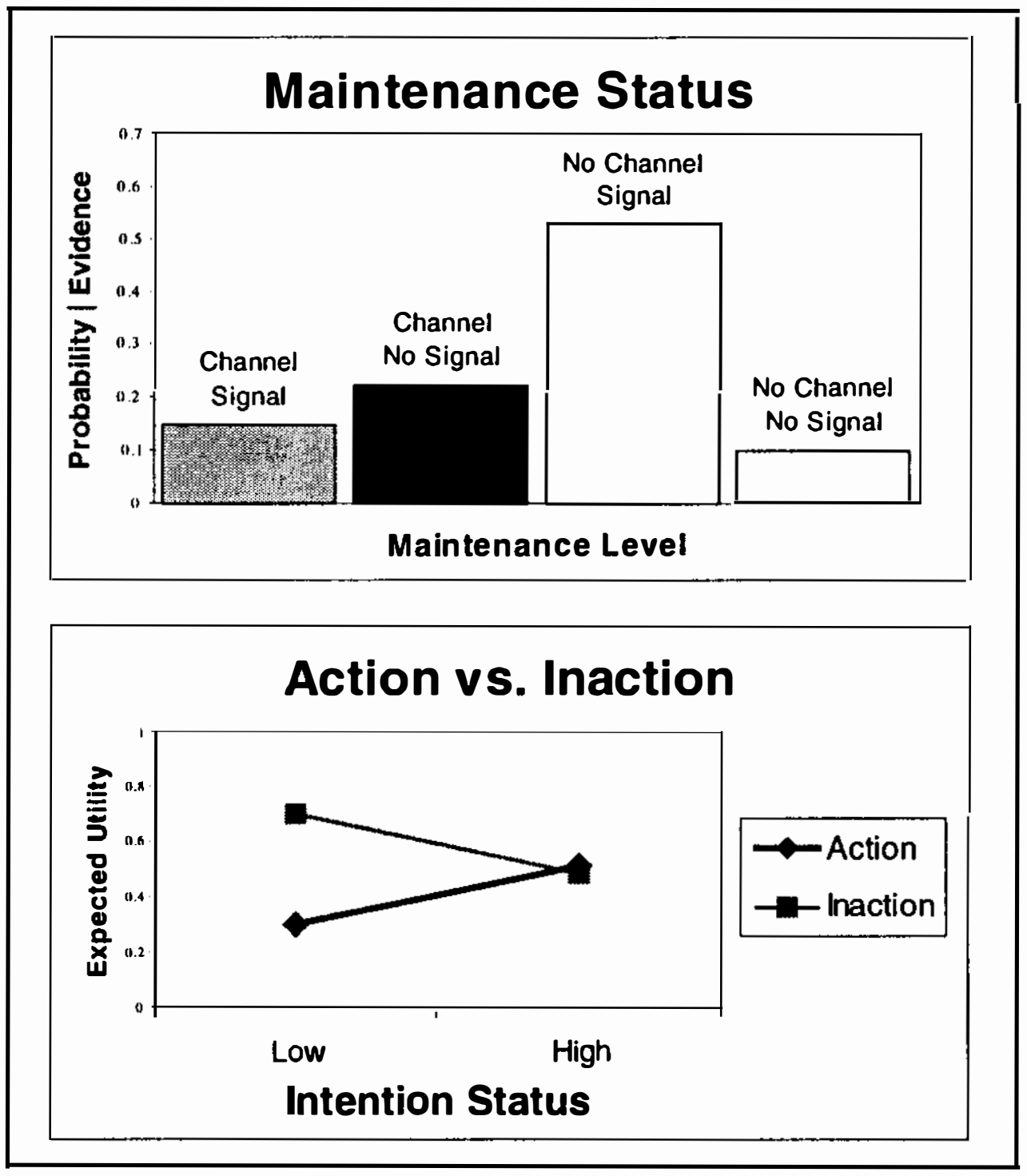

Figure 8. Change in expected utility as Presenter tries to distinguish overheard from intended speech.

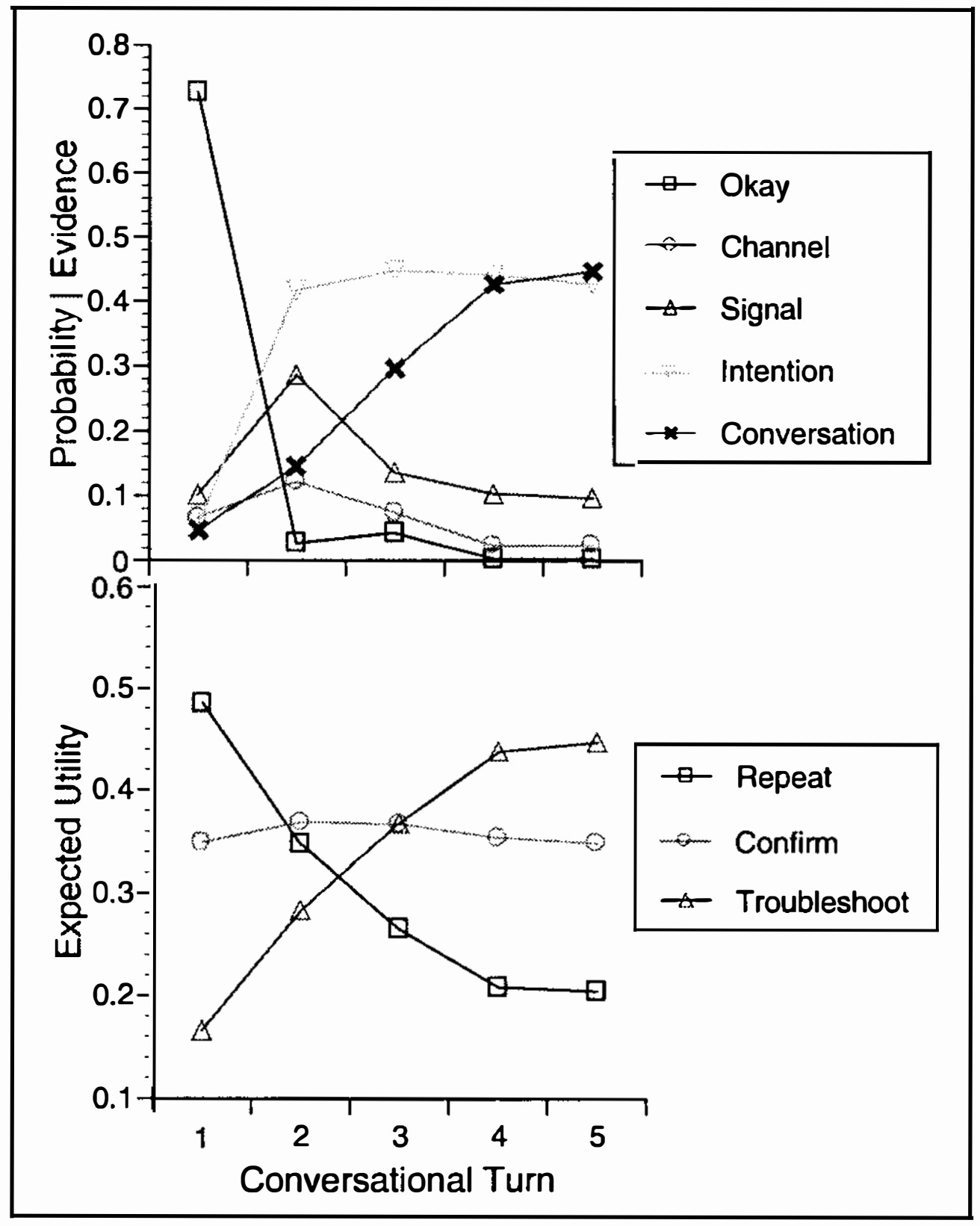

Figure 9. Two graphs demonstrating how grounding strategies adapt to the Grounding Status.

## 3.3 ADAPTING STRATEGIES OVER TIME

To illustrate how Quartet adapts its strategies over time as mutual understanding fluctuates, consider the case where Presenter is faced with the task of deciding between doing a requested service and resorting to only three repair strategies: ASK REPEAT, CONFIRM, and TROUBLESHOOT. In ASK REPEAT, the system asks the user to repeat the previous utterance while communicating its uncertainties; in CONFIRM, the system asks the user to confirm its best guess of the goal or to elaborate on the previous user utterance; and finally, in TROUBLESHOOT, the system steps outside of the dialog and uses VOI analysis to recommend procedures to improve mutual understanding at the appropriate level. TROUBLESHOOT may also include the termination of the joint activity itself.

The graphs in Figure 9 display a conversation in which the user repeatedly corrects Presenter for engaging in the wrong action. As the top panel shows, after being corrected in the first turn, the probability that the "Grounding Status" is OKAY drops down while the probability of failure at different levels gradually moves up. Since correction implies that the system is guessing the wrong intentions of the user, the likelihood of INTENTION FAILURE exceeds everything else. Furthermore, since CONFIRM is best suited for dealing with intention level failures, it has the highest expected utility. The TROUBLESHOOT strategy is appropriately low in starting off a conversation.

As the conversation proceeds and the user continues to correct Presenter, despite receiving high confidence signals from the ASR engine and natural language parser, the probability of a maintenance level problem decreases, as shown in the dwindling likelihood of SIGNAL and CHANNEL FAILURE in Figure 9. Notice that the expected utility graph mirrors this change with a fall in the expected utility of ASK REPEAT, a strategy that is best suited for maintenance level problems.

Since CONFIRM has the highest expected utility, Presenter continues to ask for confirmations while the user rejects the confirmed action. Eventually as dialog record continues to accumulate more intention errors, and the number of turns increases, the likelihood of a CONVERSATION failure quickly grows. Since the counterpart strategy to CONVERSATION failure is TROUBLESHOOT, its expected utility begins to surpass all other repair strategies after the third turn, as intuition would suggest. If VOI troubleshooting fails, the system have no other recourse than to apologize and terminate the joint activity.

# 4 CONCLUSIONS

We have outlined an approach to continuous spoken dialog centering on an architecture called Quartet. Quartet employs models that represent and reason about key uncertainties at four levels of mutual understanding. In contrast to the majority of spoken dialog systems that focus almost entirely on the intention level, Quartet reasons explicitly about uncertainties that can lead to communication failures at any of the levels. The approach facilitates task independent, multimodal grounding through the use of Bayesian networks, expected utility, and value-of-information analyses. Research in psychology and linguistics informs the overall modularity and interdependence of the levels considered in the system, as well as the variables and structure of the Bayesian models employed at different levels.

We have also reviewed several examples of Quartet in action, highlighting the potential for endowing spoken dialog systems with new forms of robustness via the introduction of grounding machinery and procedures. Furthermore, we have illustrated how Quartet can employ reasoning about mutual understanding at multiple levels to discriminate overheard utterances from those directed at the system, and to adapt its repair strategies to meet the grounding situation of the conversation as it progresses over time.

Rather than writing ad hoc policies to deal with failures at multiple, interdependent levels of conversation, or waiting for enhancements in the precision of dialog components, we have focused on representations, inference procedures, and decision strategies for designing spoken dialog systems with the ability to manage uncertainties through grounding. These representations and procedures explicitly consider the limitations of the components in deliberating about the best way to achieve mutual understanding. In the near term, we expect this research leading to dialog systems that are more robust to local communication failures by employing flexible grounding procedures to handle misunderstanding in a natural manner. In the long term, we foresee the evolution of this work leading eventually to fluid conversations between people and computers.

## References

Ben-Bassat, M. & Teeni, D. 1984. Human-oriented information acquisition in sequential pattern classification: Part 1 – single members. *IEEE Transactions on Systems, Man, & Cybernetics.* 14: 131-138.

Brennan, S.A. & Hulteen, E. 1995. Interaction and feedback in a spoken language system: A theoretical framework. *Knowledge-Based Systems* 8: 143-151.

Clark, H.H. 1996. *Using Language.* Cambridge University Press.

Clark, H.H. & Brennan, S.A.. 1991. Grounding in communication. In *Perspectives on Socially Shared Cognition*, 127-149. APA Books

Clark, H.H. & Schaefer, E.F. 1987. Collaborating on contributions to conversations. *Language and Cognitive Processes* 2(1): 19-41.

Clark, H.H. & Schaefer, E.F. 1989. Contributing to discourse. *Cognitive Science* 13: 259-294.

Clark, H.H. & Wilkes-Gibbs, D. 1990. Referring as a collaborative process. In *Intentions in Communication*, 463-493. MIT Press.

Cohen, P.R. & Levesque, H.J. 1991. Teamwork. *Nous*, 25(4): 487-512.

Cohen, P.R. & Levesque, H.J.. 1994. Preliminaries to a collaborative model of dialogue. *Speech Communication* 15: 265-274.

Conati, C., Gertner, A., VanLehn, K., & Druzdzel, M., 1997. Online student modeling for coached problem solving using Bayesian networks. *Proc. of the Sixth International Conference on User Modeling*, 231-242. Springer-Verlag.

Dagum, P., Galper, A., & Horvitz, E. 1992. Dynamic network models for forecasting. In *Proceedings of the Eighth Workshop on Uncertainty in Artificial Intelligence*, 41-48.

Dillenbourg, P., Traum, D. & Schneider, D. 1996. Grounding in multi-modal task-oriented collaboration. *Proc. of the EuroAI & Education Conference.*

Goodwin, C. 1986. Between and within: Alternative sequential treatments of continuers and assessments. *Human Studies* 9:205-217.

Gorry, G.A. & Barnett, G.O. 1968. Experience with a model of sequential diagnosis. *Computers and Biomedical Research* 1:490-507.

Heckerman, D.E., Horvitz, E., & Nathwani, B.N. 1992. Toward normative expert systems: Part I. *The Pathfinder project. Methods of Information in Medicine* 31:90-105.

Hirst, G., McRoy, S., Heeman, P., Edmonds, P., & Horton, D. 1994. Repairing conversational misunderstandings and non-understandings. *Speech Communication* 15: 213-229.

Horvitz, E. 1997. Agents with beliefs: Reflections on Bayesian methods for user modeling. *Proc. of the Sixth International Conference on User Modeling*, 441-442. Springer-Verlag.

Horvitz, E., Breese, J., & Henrion, M. 1988. Decision theory in expert systems and artificial intelligence. In *Journal of Approximate Reasoning, Special Issue on Uncertainty in Artificial Intelligence*, 2:247-302.

Horvitz, E., Breese, J., Heckerman, D., Hovel, D., & Rommelse, D., 1998. The Lumière project: Bayesian user modeling for inferring the goals and needs of software users. *Fourteenth Conference on Uncertainty in Artificial Intelligence,* 256-265. Morgan Kaufmann.

Horvitz, E. & Paek, T. 1999. A computational architecture for conversation. *Proc. of the Seventh International Conference on User Modeling*, 201-210. Springer Wien.

Horvitz, E. & Paek, T. 2000. Harnessing representations of goals and actions to enhance the robustness of speech recognition. Submitted for publication.

Huang, X., Acero, A., Alleva, F., Hwang, M., Jiang, L., & Mahajan, M. (1995). Microsoft Windows highly intelligent speech recognizer: WHISPER. In *Proc. of ICASSP.* IEEE.

Kanazawa, K., & Dean, T. 1989. A model for projection and action. In *Proceedings of the Eleventh IJCAI.* AAAI/IJCAI.

Kanazawa, K., Koller, D., & Russell, S. 1995. Stochastic simulation algorithm for dynamic probabilistic networks. In *Proceedings of the Eleventh Annual Conference on Uncertainty in AI*, 346-351.

Nicholson, A., & Brady, J. 1994. Dynamic belief networks for discrete monitoring. *IEEE Transactions on Systems, Man, and Cybernetics*, 24(11): 1593-1610.

Paek, T. & Horvitz, E. 1999. Uncertainty, utility, and misunderstanding. *AAAI Fall Symposium on Psychological Models of Communication*, North Falmouth, MA, November 5-7, 85-92.

Paek, T. & Horvitz, E. 2000. Grounding criterion: Toward a formal theory of grounding. *MSR Technical Report*, MSR-TR-2000-40. Submitted for publication.

Raiffa, H. 1968. *Decision Analysis: Introductory Lectures on Choice Under Uncertainty.* Addison-Wesley, MA.

Richardson, S. 1994. Bootstrapping statistical processing into a rule-based natural language parser. Unpublished *MSR Technical Report.*

Toyama, K. 1998. Prolegomena for robust face tracking. *MSR Technical Report*, MSR-TR-98-65.

Traum, D. 1994. A computational theory of grounding in natural language conversation, Ph.D. thesis, Dept. of Computer Science, Rochester.

Traum, D. & Dillenbourg, P. 1996. Miscommunication in multi-modal collaboration. *AAAI Workshop on Detecting, Repairing, And Preventing Human--Machine Miscommunication*, 37-46.

Traum, D. & Dillenbourg, P. 1998. Towards a normative model of grounding in collaboration. *ESSLLI-98 Workshop on Mutual Knowledge, Common Ground and Public Information.*